\documentclass[runningheads]{llncs}
\usepackage{graphicx}
\usepackage{url}
\usepackage{array}
\usepackage{amssymb}
\usepackage{multirow}
\usepackage{subfigure} 
\usepackage{caption}
\usepackage{booktabs}
\usepackage{xcolor}
\usepackage{float}
\usepackage[colorlinks,linkcolor=red]{hyperref}

\begin{document}
\title{A Temporal Attentive Approach for Video-Based Pedestrian Attribute Recognition}
%
%
 \author{Zhiyuan~Chen\thanks{Student first author.}  \and
 Annan~Li \and
 Yunhong~Wang}

%
%
\institute{School of Computer Science and Engineering, Beihang University, Beijing, China
\email{{\{dechen, liannan, yhwang\}@buaa.edu.cn}}}
%
\maketitle              
\begin{abstract}
In this paper, we first tackle the problem of pedestrian attribute recognition by video-based approach. 
The challenge mainly lies in spatial and temporal modeling and how to integrating them for effective and dynamic pedestrian representation.
To solve this problem, a novel multi-task model based on the conventional neural network and temporal attention strategy is proposed.
Since publicly available dataset is rare, two new large-scale video datasets with expanded attribute definition are presented, on which the effectiveness of both video-based pedestrian attribute recognition methods and the proposed new network architecture is well demonstrated.
The two datasets are published on \url{http://irip.buaa.edu.cn/mars_duke_attributes/index.html}.

\end{abstract}
\section{Introduction}
\label{sec:intro}
Pedestrian attribute, such as gender, age and clothing characteristics, has drawn a great attention recently due to its wide range of applications in intelligent surveillance system. It can be used for retrieving pedestrian and assisting other computer vision tasks, such as human detection~\cite{cvpr2015det}, person re-identification~\cite{bmvc2012reid,tcsvt2015reid,icb2015reid,icpr2016reid,tpami2018reid,cvpr2018reid,chang2018multi} etc.

In the past years, a lot of effort has been made to pedestrian attribute recognition. Layne et al.~\cite{bmvc2012reid}, Deng et al.~\cite{mm2014attr} and Li et al.~\cite{tcsvt2015reid} use support vector machines to recognize pedestrian attribute, while AdaBoost is utilized by Zhu et al.~\cite{iccvw2013attr}. Recently, Convolutional Neural Networks (CNN) have been adopted. Sudowe et al.~\cite{iccvw2015attr} propose a jointly-trained holistic CNN model, while Li et al.~\cite{acpr2015attr} investigate CNN for both individual and group attributes. Liu et al.~\cite{iccv2017attr_atten} introduce attention model to CNN-based pedestrian attribute recognition. Wang et al.~\cite{iccv2017attr_joint}, use recurrent learning for modeling the attribute correlations. Zhao et al.~\cite{ijcai2018attr} further improve such approach by analyzing intra-group and inter-group correlations. Since clothing attribute is highly relevant to spatial location, Zhang et al.~\cite{zhang2014panda} and Li et al.~\cite{icme2018attr} use pose estimation for assistance.

\begin{figure}[t]
	\centering
	\includegraphics[width=0.99\textwidth]{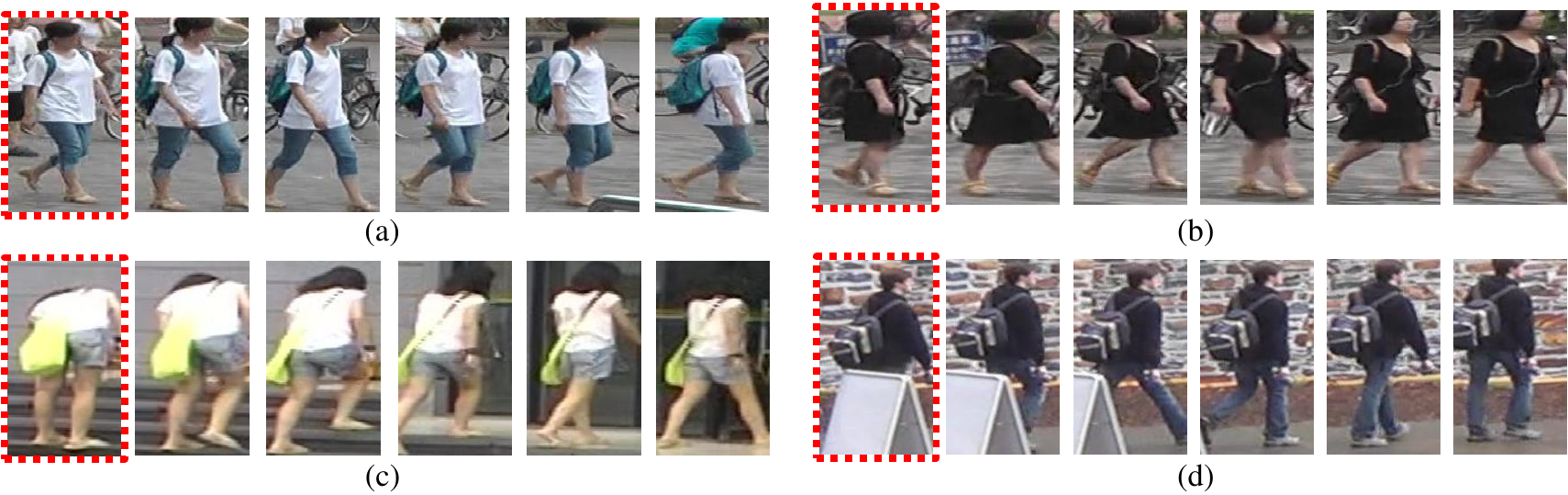}
	\vspace{-2mm}
	\caption{Comparison between image and video based  pedestrian attribute recognition. (a) Backpack is invisible in the frontal view, but can be clearly observed in other image of a sequence. (b) For a woman in dark color, it is difficult to classify whether she is wearing a skirt or shorts using one single image of limited resolution. However, the difficulty can be mitigated with the swinging of skirt. (c) The act of bowing head may makes it impossible to recognize the hair length attribute of the pedestrain from one single image, while video data contains richer motion information. (d) Quality problems like occlusion may highly affected the recognition progress of some specific attributes in still-images.}
	\label{fig:vid}
\end{figure}

Although demonstrated good performance, the above-mentioned methods are all based on static image. They are trained and evaluated on datasets with only one image per instance~\cite{mm2014attr,arXiv2016RAP,iccv2015market1501,Cheng2018Pedestrian,Sarfraz2017Deep,he2017adaptively,sun2018unified}. However, in a real-world surveillance scenario, consecutive image sequence is available. As can be seen from Figure~\ref{fig:vid}~(a), a single shot of pedestrian (dashed rectangle) is not necessarily the most representative one for a specific attribute. Besides that sequential data can also provide strong temporal cues (see Figure~\ref{fig:vid}~(b)), which are overlook in existing image-based approaches. What's more, as shown in Figure~\ref{fig:vid}~(c) and Figure~\ref{fig:vid}~(d), video data shows clear superiority in handling some special cases and quality problems. It is reasonable that pedestrian attribute recognition should be tackled by video-based approach.

In this paper, a novel deep learning approach for video-based pedestrian attribute recognition is proposed. To our knowledge, it is the first one tackling pedestrian attribute recognition by video. Lack of data is the possible reason why existing approaches are limited to static image. To address this problem, we annotate two large-scale datasets of pedestrian image sequences with rich attribute. Experimental results clearly demonstrate that the proposed approach is very effective. Detailed contributions of this paper include:
\begin{itemize}
	\item Two large-scale pedestrian video datasets with rich attribute annotation are presented.
	\item A novel multi-task model based conventional neural network and temporal attention strategy is proposed for pedestrian attribute recognition.
	\item Extensive experiments are conducted and the results clearly show the superiority of video-based pedestrian attribute recognition.
\end{itemize}

The rest of this paper is organized as follows. The next section describes the annotated datasets. Then, Section~\ref{sec:method} introduces the proposed video-based pedestrian attribute recognition approach. Experimental results are shown in Section~\ref{sec:exp} and conclusion is drawn in Section~\ref{sec:conclude}.

\begin{figure*}[t]
	\centering
	\includegraphics[width=0.95\textwidth]{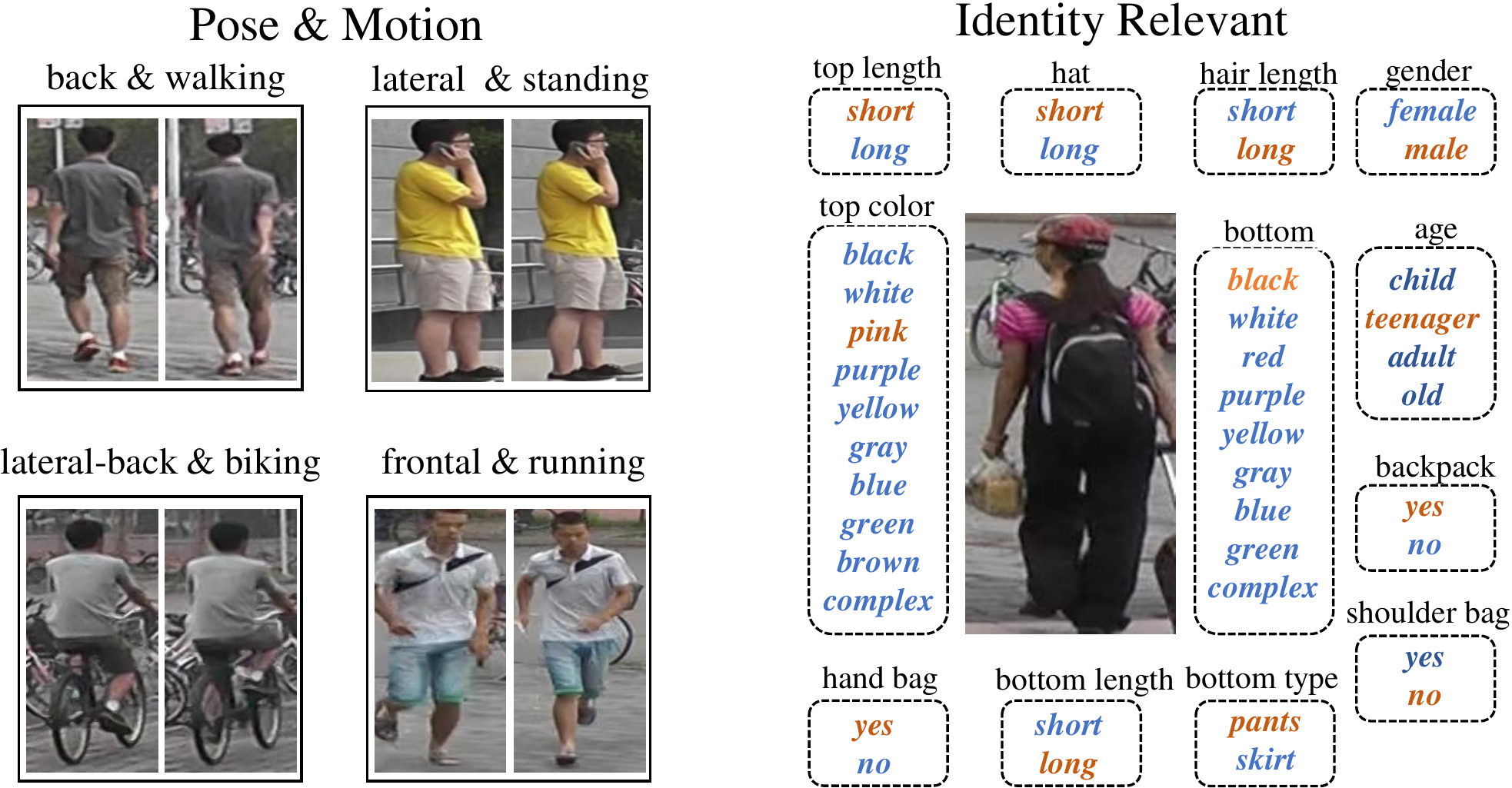}
	\vspace{-3mm}
	\caption{Exemplar attribute annotation.}
	\label{fig:attr}
\end{figure*}
\section{Dataset}
\label{sec:data}

\begin{figure}[h]
	\centering
	\includegraphics[width=0.9\textwidth]{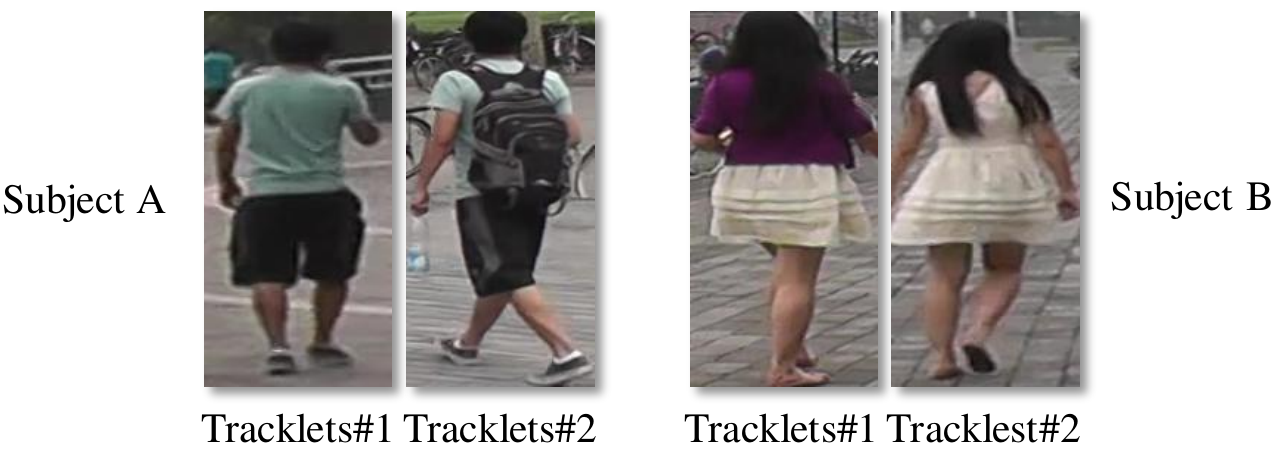}
	\vspace{-2mm}
	\caption{Examples of attribute change over time in MARS. A man puts on a backpack (left) and  woman takes off her cardigan (right). Using the attributes annotated on a single image sequence for all the video instances of a pedestrian may incorporate errors.}
	\label{fig:change}
\end{figure}

Existing datasets for pedestrian attribute recognition such as PETA~\cite{mm2014attr}, RAP~\cite{arXiv2016RAP} and Market-1501~\cite{iccv2015market1501,lin2017improving} are mainly image-based. Fortunately, with the progress in video-based person re-identification (ReID), large-scale dataset of pedestrian image sequence becomes available. The Motion Analysis and Re-identification Set (MARS)~\cite{eccv2016mars} and DukeMTMC-VideoReID~\cite{wu2018exploit} are newly released datasets, MARS consists of 20,478 tracklets from 1,261 people captured by six cameras, while DukeMTMC-VideoReID dataset contains 4,832 tracklets from 1,402 different pedestrians captured by eight cameras. MARS is an extension of Market-1501, they share the same identity, DukeMTMC-VideoReID is also an extension of DukeMTMC-ReID which also follows the same identity rules. Although Lin et al.~\cite{lin2017improving} provides identity-level attribute annotation for Market-1501 and DukeMTMC-ReID, these annotations cannot be directly adopted to MARS and DukeMTMC-VideoReID for two reasons: Firstly, instance correspondences between the imaged-based dataset and the video-based dataset are not one-to-one; Secondly, as can be seen from Figure~\ref{fig:change}, due to some temporal changes, even for the same people in different tracklets, some attribute appears while some attribute disappears. Therefore the identity-level annotation of Market-1501 and DukeMTMC-ReID is inaccurate for MARS and DukeMTMC-VideoReID.

To address the above-mentioned problem, we build new datasets by re-annotating MARS and DukeMTMC-VideoReID using an extended attribute definition based on Lin et al.~\cite{lin2017improving}. As shown in Figure~\ref{fig:attr}, there are 16 kinds of attributes are labeled for each tracklets in MARS dataset: motion (walking, standing, running, biking, various), pose (frontal, lateral-frontal, lateral, lateral-back, back, various), gender (male, female), length of hair (long, short), length of tops/sleeve (long, short), length of bottoms (long, short), type of bottoms (pants, dress), wearing hat (yes, no), carrying shoulder bag (yes, no), carrying backpack (yes, no), carrying handbag (yes, no), nine bottom colors (black, white, red, purple, yellow, gray, blue, green, complex), ten top colors (black, white, pink, purple, yellow, gray, blue, green, brown, complex) and four kinds of ages (child, teenager, adult, old) which results in a total attribute number of 52. The DukeMTMC-VideoReID dataset is also re-annotated with the same expanded attribute definition rule.

The attributes can be divided into two categories: identity-relevant and behavior-relevant. Prior arts only focus on the former one since their main purpose is retrieving people from surveillance video. However, as can be seen from Figure~\ref{fig:attr} (left column), behavior-relevant factors can greatly influence the appearance of a pedestrian. We argue that identifying such attributes is not only useful for comprehensive pedestrian understanding but also beneficial for ID-relevant attribute recognition itself. Because excluding the distraction caused by behavior can improve the focus on salient frame that containing identical attribute feature.



\section{Approach}
\label{sec:method}

In this section, we first describe the overall architecture of our pedestrian attribute recognition network. Then we give a detailed introduction of the temporal attention strategy of the architecture.
\subsection{Network Architecture}
\label{subsec:overview}

\begin{figure}[t]
	\centering
	\includegraphics[width=0.99\textwidth]{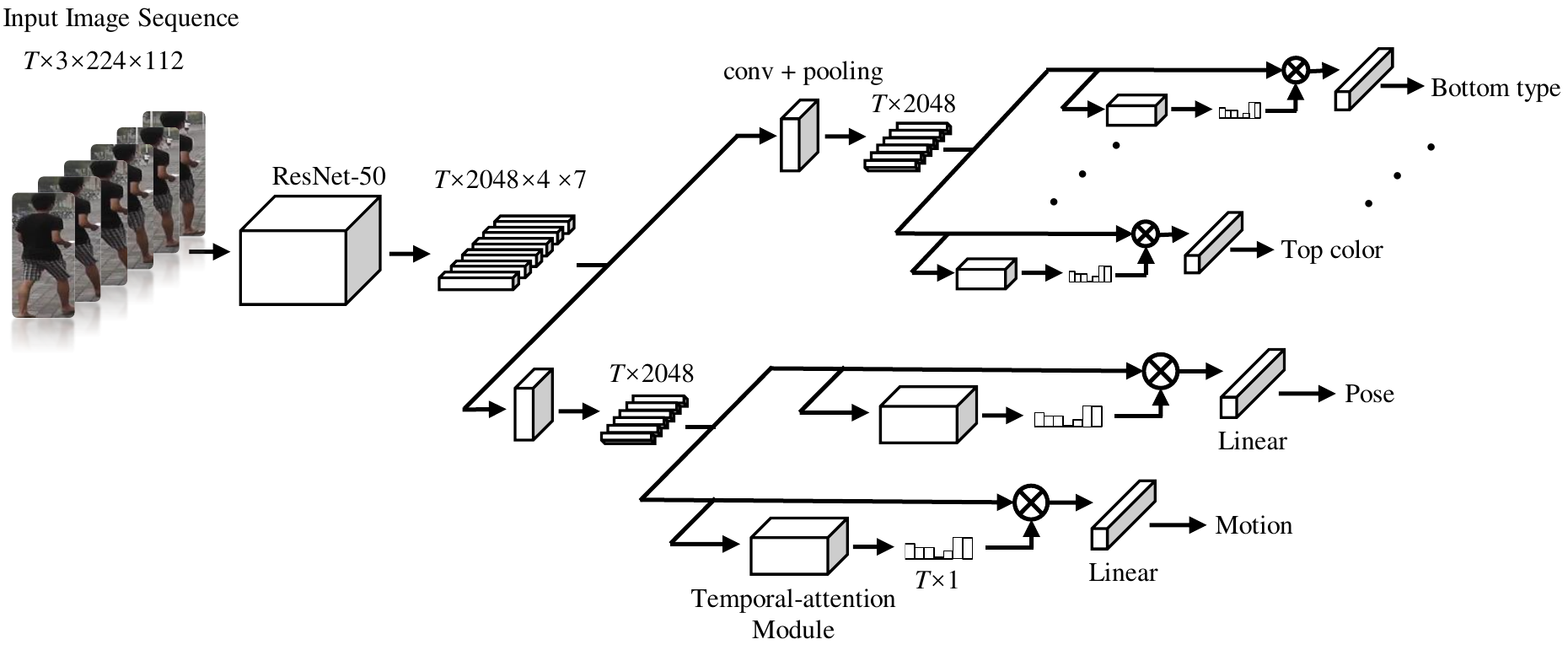}
	\vspace{1mm}
	\caption{Overall architecture of the proposed model. For an input pedestrian image sequence, frame-level spatial feature is first extracted by using the ResNet-50 network, then the spatial features is processed by two separated channels, one is the motion \& pose attribute recognition channel which processed the spatial features by an unique convolution + pooling unit then put pooled features into single attribute recognition classifiers, the other channel shares similar architecture with the first one but takes the responsibility for recognizing ID-relevant attributes such as top color, gender or bottom type. Each single attribute recognition classifier contains a temporal-attention module which is used to generate temporal attention for the single attribute recognition process. }
	\label{fig:flowchart}
\end{figure}

The overall architecture of our proposed model is illustrated in Figure~\ref{fig:flowchart}. At the beginning of the network, we choose ResNet-50~\cite{cvpr2016resnet} as the backbone model, and the outputs of last flatten layer are used as the frame-level spatial feature, then the network is separated into two channels: i.e. the motion \& pose channel and the ID-relevant channel respectively. The reason why we separate the classifiers into two channels is that motion \& pose attributes are ID-irrelevant and its classifier would focus on different parts of the spatial features compared with the id-relevant attributes, so directly sharing the same spatial features among all the id-irrelevant and id-relevant attribute classifiers will lead to a feature-compete situation, which means both the id-irrelevant classifiers and id-relevant classifiers would restrain each other in the training progress. The effectiveness of this separation will be validated in the experiments.

Let $I=\{I_{1},I_{2},...,I_{n}\}$ be an input image sequence or a tracklet, where $n$, $w$ and $h$ are the frame number, image width and height respectively, and we choose $n = 6, w = 112, h = 224$ in practice. Using the spatial feature extractor Resnet-50, each frame is represented by a tensor sized $2048\times 4\times 7$. Then the spatial feature vector is respectively processed by the convolution + pooling units in the two channels. Consequently, the $n\times 3\times w\times h$ tensor is converted into a two-dimensional matrix $S=\{S_1, S_2,...,S_n\}, S\in\mathbb{R}^{n\times 2048}$.

Then the pooled spatial feature vector is processed by the attribute classifiers. Firstly the temporal-attention module in each attribute classifier would take the spatial feature vector as input and generate a temporal attention vector $A$ sized $n\times 1$ which represents the importance of each frame in recognizing the specific attribute. Then the temporal attention vector is used to weight the spatial feature of each frame, and a final feature vector of the image sequence for recognizing a specific attribute would be generated $F = A^{T} \times S$. Lastly the final feature vector would be fed into fully connected layer to achieve the attribute classification results.

We evaluate the influence of both the separated channel strategy and the temporal attention strategy in term of attributes recognition accuracy in Section~\ref{sec:exp}, and the results shows that the corresponding strategies are the best choices for video-based pedestrian attributes recognition.


\subsection{Temporal-attention Strategy}
\label{subsec:temporalattention}

Although ResNet-50 is able to capture effective spatial information from each single frame, however, we find that the importance of each frame in recognizing different attributes may vary. In other words, some frames may be greatly helpful in recognizing one attribute but may be harmful to another. As can be seen from Figure~\ref{fig:vid} and Figure~\ref{fig:attr}, the recognition of different attributes may rely on different key frames, therefore each single attribute classifier is equipped with a temporal-attention module, which is also helpful in reducing the negative influence introduced by sharing the same spatial feature vector among the classifiers.
 
As shown in Figure~\ref{fig:attentions}, by applying independent temporal-attention modules on different classifiers, various temporal attention vectors would be generated to be adaptive to the attribute classifier it serves. The superiority of this temporal attention strategy will be presented in Section~\ref{subsec:ablation-exp}.
\section{Experiments}
\label{sec:exp}
In this section, firstly we give a brief description about the train/test set partition of the annotated MARS and DukeMTMC-VideoReID datasets as well as some training/testing settings in the experiments. Then we compare the performance of the proposed method with the image-based method as well as other video analysis model such as 3DCNN~\cite{ji20133d} and CNN-RNN model~\cite{mclaughlin2016recurrent}, which demonstrates the superiority of our multi-task architecture in video-based pedestrian attribute recognition. Lastly ablation study shows the effectiveness of the separated channel strategy and temporal-attention strategy.

\subsection{Settings}
We follow the original train/test set partition rule of MARS~\cite{eccv2016mars} and DukeMTMC-VideoReID~\cite{wu2018exploit}. The training set of MARS consists of 8,298 tracklets from 625 people, while the rest 8,062 tracklets corresponding to 626 pedestrians make up the test set, the average frame number among these tracklets is 60. DukeMTMC-VideoReID has a smaller tracklets number but a larger average frame number which is 169. Both two datasets shares no identity in its train and test sets due to the peculiarity of the person re-identification task.
\begin{figure}[H]
	\centering
	%
	\includegraphics[width=\textwidth]{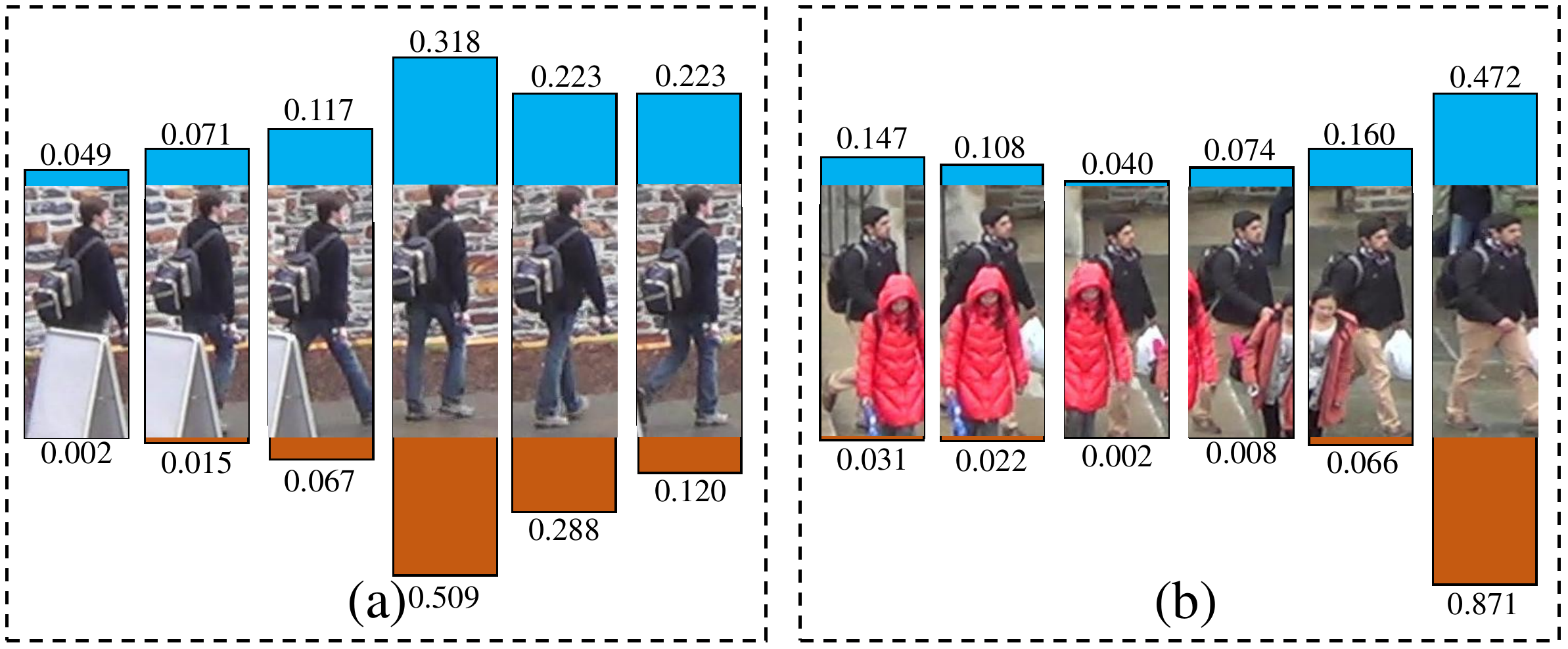}
	\includegraphics[width=\textwidth]{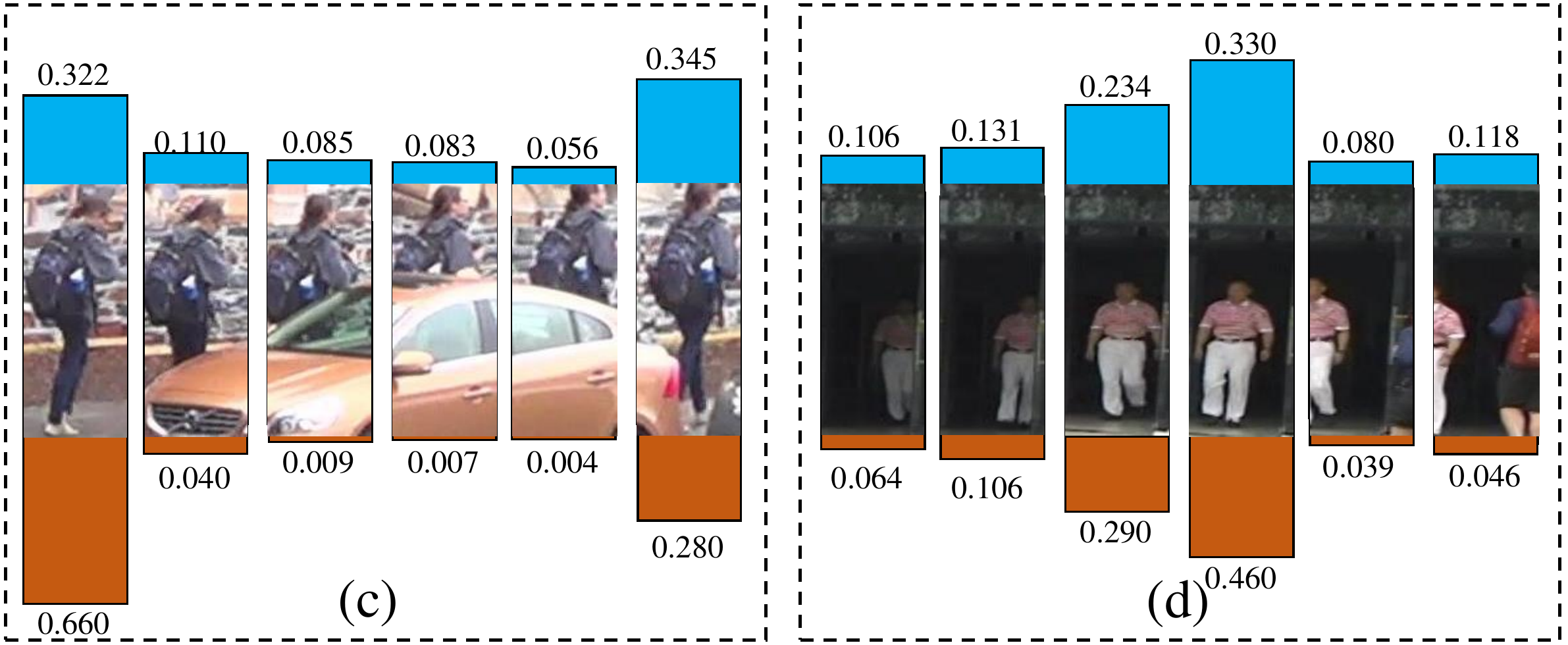}
	\includegraphics[width=\textwidth]{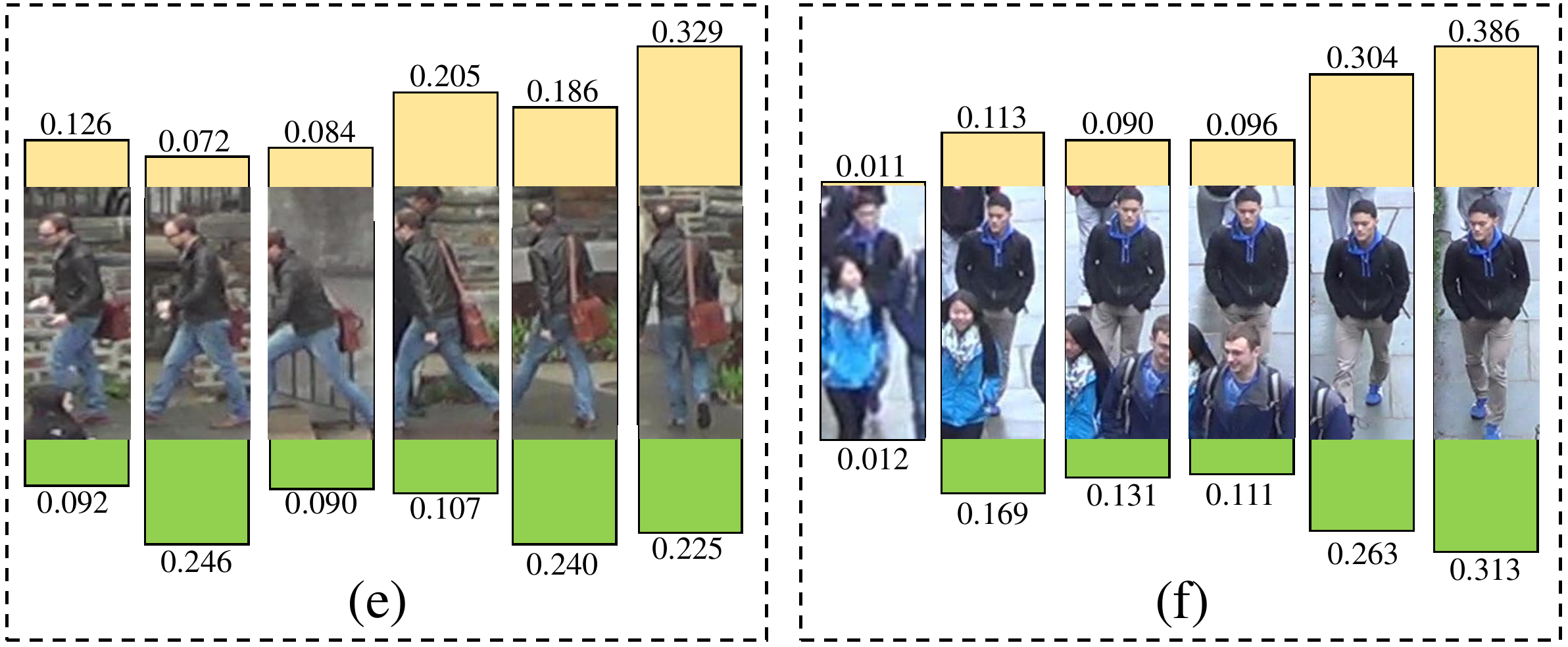}
	\includegraphics[width=\textwidth]{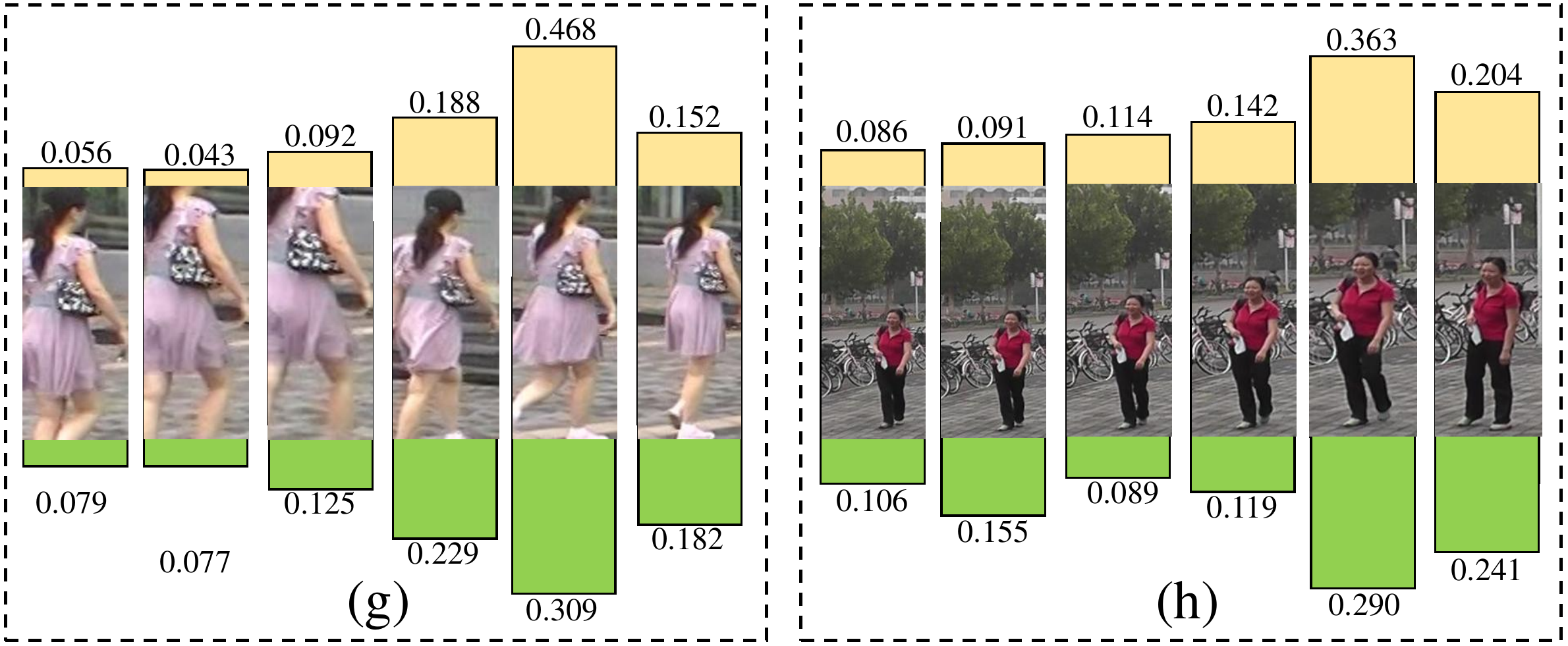}
	\caption{Examples of temporal attention. The blue/brown/yellow/green bars represent the temporal attention vector generated by the \emph{backpack}/\emph{bottom color}/\emph{shoulder bag}/\emph{boots} attribute classifier respectively. }
	\label{fig:attentions}

\end{figure}
In the training progress, to form a training batch, firstly we random select $K = 64$ tracklets from the train set, then $n = 6$ frames is randomly sampled from each tracklets, so each training batch is formed by $K \times n$ frames, random sampling strategy is more suitable to temporal attention models for it increases the variance among the sampled frames compared with consecutive sampling strategy. In the testing process, for each testing tracklets which contains $F$ frames, these frames would be randomly split into $\lfloor \frac{F}{n} \rfloor$ groups, and the testing attribute prediction result is the average prediction result among these groups. Cross Entropy Loss is chosen as the loss function and Adam with a learning rate 0.0003 is selected as the optimizer in training.
\begin{table}[t]
	\caption{Comparisons of recognition accuracy and F1 measure on MARS datasets(\%).}
	\centering
	\begin{tabular}{|p{2cm}|c|c|c|c|c|c|c|c|}
		\hline
		\multirow{3}{*}{Attribute}&\multicolumn{2}{c|}{Image-based}&\multicolumn{2}{c|}{\multirow{2}{*}{3DCNN}}&\multicolumn{2}{c|}{\multirow{2}{*}{CNN-RNN}}&\multicolumn{2}{c|}{\multirow{2}{*}{ours}}\\
		& \multicolumn{2}{c|}{Baseline}&\multicolumn{2}{c|}{}&\multicolumn{2}{c|}{}&\multicolumn{2}{c|}{} \\
		\cline{2-9}
		& acc & F1 & acc & F1 & acc & F1 & acc & F1\\
		\cline{1-9}
		motion&91.08&39.39&90.34&33.64&\textbf{\color{blue}{92.12}}&\textbf{\color{blue}{43.92}}&\textbf{\color{blue}{92.12}}&43.69\\
		pose&72.03&56.91&62.51&47.69&72.40&58.36&\textbf{\color{blue}{73.65}}&\textbf{\color{blue}{61.36}}\\
		top color&\textbf{\color{blue}{74.73}}&\textbf{\color{blue}{72.72}}&68.04&65.63&71.90&69.28&73.43&71.44\\
		bottom color&68.27&\textbf{\color{blue}{44.63}}&65.44&40.39&65.77&39.68&\textbf{\color{blue}{69.45}}&43.98\\
		age&83.44&38.87&81.70&36.22&84.28&39.93&\textbf{\color{blue}{84.71}}&\textbf{\color{blue}{40.21}}\\
		top length&94.21&58.72&93.63&56.37&\textbf{\color{blue}{94.60}}&65.18&94.47&\textbf{\color{blue}{71.61}}\\
		bottom length&92.69&92.29&89.96&89.35&93.70&93.33&\textbf{\color{blue}{94.22}}&\textbf{\color{blue}{93.90}}\\
		shoulder bag&80.39&72.57&71.82&61.30&82.70&75.89&\textbf{\color{blue}{83.48}}&\textbf{\color{blue}{76.08}}\\
		backpack&89.37&85.95&82.60&76.58&90.18&87.17&\textbf{\color{blue}{90.59}}&\textbf{\color{blue}{87.62}}\\
		hat&96.91&57.57&96.53&57.69&\textbf{\color{blue}{97.90}}&77.74&97.51&\textbf{\color{blue}{77.84}}\\
		hand bag&85.71&62.82&83.88&59.90&\textbf{\color{blue}{88.07}}&71.68&87.61&\textbf{\color{blue}{73.55}}\\
		hair&88.61&86.91&85.12&82.77&88.78&87.11&\textbf{\color{blue}{89.54}}&\textbf{\color{blue}{88.17}}\\
		gender&91.32&90.89&86.49&85.75&92.77&92.44&\textbf{\color{blue}{92.83}}&\textbf{\color{blue}{92.50}}\\
		bottom type&93.12&81.69&89.19&72.86&93.67&84.16&\textbf{\color{blue}{94.60}}&\textbf{\color{blue}{86.62}}\\
		\hline
		average.&85.85&67.28&81.95&61.87&86.35&70.42&\textbf{\color{blue}{87.01}}&\textbf{\color{blue}{72.04}}\\
		\hline
	\end{tabular}
	\label{table:attributemars}
\end{table}
\begin{table}[t]
	\caption{Comparisons of recognition accuracy and F1 measure on DukeMTMC-VideoReID dataset(\%).}
	\footnotesize
	\centering
	\begin{tabular}{|p{2cm}|c|c|c|c|c|c|c|c|}
		\hline
		\multirow{3}{*}{Attributes}&\multicolumn{2}{c|}{Image-based}&\multicolumn{2}{c|}{\multirow{2}{*}{3DCNN}}&\multicolumn{2}{c|}{\multirow{2}{*}{CNN-RNN}}&\multicolumn{2}{c|}{\multirow{2}{*}{ours}}\\
		& \multicolumn{2}{c|}{Baseline}&\multicolumn{2}{c|}{}&\multicolumn{2}{c|}{}&\multicolumn{2}{c|}{} \\
		\cline{2-9}
		& acc & F1 & acc & F1 & acc & F1 & acc & F1\\
		\cline{1-9}
		motion&97.65&19.76&97.68&21.37&\textbf{\color{blue}{97.76}}&26.65&97.65&\textbf{\color{blue}{27.68}}\\
		pose&72.46&62.63&69.46&59.95&74.36&66.26&\textbf{\color{blue}{75.31}}&\textbf{\color{blue}{67.73}}\\
		backpack&87.41&86.12&81.05&77.59&89.78&87.95&\textbf{\color{blue}{90.05}}&\textbf{\color{blue}{88.37}}\\
		shoulder bag&86.15&\textbf{\color{blue}{77.90}}&83.14&64.28&87.35&75.33&\textbf{\color{blue}{87.88}}&76.47\\
		hand bag&\textbf{\color{blue}{94.95}}&56.34&94.34&51.09&94.34&57.82&94.42&\textbf{\color{blue}{64.67}}\\
		boots&94.12&92.57&83.59&78.97&94.72&93.25&\textbf{\color{blue}{94.95}}&\textbf{\color{blue}{93.52}}\\
		gender&89.57&89.49&82.49&82.47&90.35&90.28&\textbf{\color{blue}{90.85}}&\textbf{\color{blue}{90.78}}\\
		hat&93.02&88.26&87.54&76.12&93.32&88.45&\textbf{\color{blue}{93.73}}&\textbf{\color{blue}{89.41}}\\
		shoes color&\textbf{\color{blue}{93.32}}&83.35&88.07&69.76&93.05&84.65&93.13&\textbf{\color{blue}{85.18}}\\
		top length&91.54&78.10&89.14&69.28&92.25&80.06&\textbf{\color{blue}{92.52}}&\textbf{\color{blue}{81.06}}\\
		bottom color&76.82&47.25&75.92&48.19&78.85&51.66&\textbf{\color{blue}{79.95}}&\textbf{\color{blue}{55.95}}\\
		top color&76.25&42.57&78.39&56.14&79.98&57.20&\textbf{\color{blue}{81.28}}&\textbf{\color{blue}{58.07}}\\
		\hline
		average.&87.77&68.70&84.24&62.93&88.84&71.63&\textbf{\color{blue}{89.31}}&\textbf{\color{blue}{73.24}}\\
		\hline
	\end{tabular}
	\label{table:attributeduke}
\end{table}

\subsection{Comparison with other Approaches}
\label{subsec:ablation-exp}

The key contribution of this work is the introduction of video-based approach to pedestrian attribute recognition. To demonstrate its superiority, an imaged-based baseline ResNet-50 model with multi-classification head trained on the frame images in the two datasets is also proposed in this paper, as shown in Table~\ref{table:attributemars} and  Table~\ref{table:attributeduke}, except for several attributes which is highly-relied on the spatial feature, the proposed model and CNN-RNN model achieves better results in most attributes, which demonstrates both the effectiveness of the proposed multi-task architecture and the superiority of video-based approach in recognizing pedestrian attribute.

We also introduce two deep-based video analysis models 3DCNN and CNN-RNN model into this video-based pedestrian attribute recognition task. As shown in the results, 3DCNN is not suitable for this task, which may imply that the 3D convolution operation would lost many important spatial clues in the tracklets. CNN-RNN model works even better than our model in recognizing the motion attribute, that's because recognizing the motion attribute relies on finding important temporal clues from the tracklets which can not be achieved by only temporal attention strategy, but our model still works better in the rest attributes, this phenomenon is consistent with the observations shown in Figure~\ref{fig:vid} and Figure~\ref{fig:attentions}, it shows that temporal feature would also cause spatial information loss, and highlighting the representativeness of key frames needs necessary spatial cues. The pedestrian attributes annotated in our dataset can be relevant to any part of the body (see Figure~\ref{fig:attr}). In other words, a region important to some attributes is not necessarily the same important to others. It is reasonable that emphasizing some specific spatial region might lead to detail loss. That is the possible explanation why temporal attention strategy outperforms RNN in recognizing most of the pedestrian attributes.

\subsection{Ablation study}
\label{subsec:ablation-exp}

\begin{table}[t]
	\caption{Ablation study on two datasets(\%).}
	\centering
	\begin{tabular}{p{6cm}|c|c|c|c}
		\toprule
		\multirow{2}{*}{Model} & \multicolumn{2}{c|}{DukeMTMC-VID} & \multicolumn{2}{c}{MARS}\\
		\cline{2-5}
		{}&aver. acc&aver. F1&aver. acc&aver. F1\\
		\cline{1-5}
		Temporal Pooling Baseline&86.25&69.80&87.86&70.36 \\
		\cline{1-5}
		Baseline. + separated channels strategy&86.84&71.38&88.97&71.70 \\
		\cline{1-5}
		Baseline. + temporal attention strategy&86.93&70.59&89.09&72.40 \\
		\cline{1-5}
		Proposed Method&\textbf{\color{blue}{87.01}}&\textbf{\color{blue}{72.04}}&\textbf{\color{blue}{89.31}}&\textbf{\color{blue}{73.24}} \\
		\bottomrule
	\end{tabular}
	\label{table:ablationstudy}
\end{table}

Since we introduce the separated channel strategy and temporal attention strategy into our multi-task architecture, a series of ablation experiments are conducted to illustrate the effectiveness of these strategies. As shown in Table~\ref{table:ablationstudy}, both the two strategies can improve the recognition performance, and the temporal attention strategy contributes more observed from their results in both two metrics, that's mainly because that the temporal attention strategy can not only pick the discriminative frames out from the input tracklets, but also can ease the feature-compete phenomenon described in Section~\ref{subsec:overview}. 

The channel separation strategy solves the feature-compete problem in physical, it physically splits the attributes into two channels, and applying separated convolution + pooling operation on the same spatial features vector, which could directly restrain the feature competition between id-relevant attributes and id-irrelevant attributes. While temporal attention strategy handles this problem latently, since each attribute classifier contains a temporal-attention module, so in the backward progress, the temporal-attention module can help smooth the backward gradients passed to bottom layers, which can also play the similar role as the Separated channel strategy. 

It can be observed from Table~\ref{table:ablationstudy}, Table~\ref{table:attributemars} and  Table~\ref{table:attributeduke} that even the video-based temporal pooling baseline can outperform the image-based method, this also illustrates the superiority of video-based approach in recognizing pedestrian attribute.  
 
\section{Conclusion}
\label{sec:conclude}

In this paper, we first study pedestrian attribute recognition with video-based approach. Two new large-scale datasets for video-based pedestrian attribute recognition is presented. We also proposed a novel multi-task architecture based on the conventional neural network and temporal attention strategy. Experiments show that video-based approach is better than image-based method in recognizing pedestrian attribute and the proposed architecture is very effective.

\section{Acknowledgment}

This work was supported by The National Key Research and Development Plan of China (Grant No.2016YFB1001002).
%
%
%
%
%
\bibliographystyle{IEEEtran}
\bibliography{paper}
\end{document}